\documentclass[conference]{IEEEtran}
\IEEEoverridecommandlockouts
\usepackage{cite}
\usepackage{amsmath,amssymb,amsfonts}
\usepackage{algorithmic}
\usepackage{graphicx}
\usepackage{textcomp}
\usepackage{xcolor}
\usepackage{multirow}
\usepackage{tabularray}
\usepackage{colortbl}
\usepackage{booktabs}
\def\BibTeX{{\rm B\kern-.05em{\sc i\kern-.025em b}\kern-.08em
    T\kern-.1667em\lower.7ex\hbox{E}\kern-.125emX}}
\begin{document}

\title{Medical Report Generation Is A Multi-label Classification Problem\thanks{$\dagger$ Equal Contribution}}

\author{\IEEEauthorblockN{
\IEEEauthorblockN{1\textsuperscript{st} Yijian Fan$^\dagger$}
\IEEEauthorblockA{\textit{Australian Artificial Intelligence Institute} \\
\textit{University of Technology Sydney}\\
Sydney, Australia \\
yijian.fan@student.uts.edu.au}
\and
\IEEEauthorblockN{1\textsuperscript{st} Zhenbang Yang$^\dagger$}
\IEEEauthorblockA{\textit{Computer Science} \\
\textit{New York University}\\
New York, United States\\
zy3101@nyu.edu}
\and
\IEEEauthorblockN{1\textsuperscript{st} Rui Liu$^\dagger$}
\IEEEauthorblockA{\textit{Australian Artificial Intelligence Institute} \\
\textit{University of Technology Sydney}\\
Sydney, Australia \\
rliu0016@gmail.com}
\and
4\textsuperscript{th} Mingjie Li$^*$}
\IEEEauthorblockA{\textit{Radiation Oncology} \\
\textit{Stanford University}\\
Palo Alto, United States \\
lmj695@stanford.edu}
\and
\IEEEauthorblockN{5\textsuperscript{th} Xiaojun Chang}
\IEEEauthorblockA{\textit{Australian Artificial Intelligence Institute} \\
\textit{University of Technology Sydney}\\
Sydney, Australia \\
xiaojun.chang@uts.edu.au}}
\maketitle

\begin{abstract}
Medical report generation is a critical task in healthcare that involves the automatic creation of detailed and accurate descriptions from medical images. Traditionally, this task has been approached as a sequence generation problem, relying on vision-and-language techniques to generate coherent and contextually relevant reports. However, in this paper, we propose a novel perspective: rethinking medical report generation as a multi-label classification problem. By framing the task this way, we leverage the radiology nodes from the commonly used knowledge graph, which can be better captured through classification techniques. To verify our argument, we introduce a novel report generation framework based on BLIP integrated with classified key nodes, which allows for effective report generation with accurate classification of multiple key aspects within the medical images. This approach not only simplifies the report generation process but also significantly enhances performance metrics. Our extensive experiments demonstrate that leveraging key nodes can achieve state-of-the-art (SOTA) performance, surpassing existing approaches across two benchmark datasets. The results underscore the potential of re-envisioning traditional tasks with innovative methodologies, paving the way for more efficient and accurate medical report generation.
\end{abstract}


\section{Introduction}

\subsection{Clinical Potential}
Medical report generation (MRG)~\cite{messina2022survey} is a crucial task in modern healthcare, involving the synthesis of detailed and precise textual descriptions from medical images such as X-rays, CT scans, and MRIs. These reports play a vital role in the clinical workflow by documenting key findings, diagnostic impressions, and recommendations for further action. They serve as a critical communication tool between radiologists, referring physicians, and other healthcare providers, ensuring that all parties have a clear understanding of the patient's condition and the necessary steps for management and treatment. The ability to generate accurate and comprehensive reports is essential for ensuring that healthcare providers can make informed decisions, track patient progress, and coordinate care effectively.

The clinical value of medical reports cannot be overstated, as they often form the basis for diagnosing diseases, planning treatments, and monitoring patient outcomes. The demand for radiology services has been increasing exponentially, driven by factors such as aging populations, advances in imaging technology, and the growing importance of imaging in preventive care and early diagnosis. Consequently, radiologists are facing a significant burden due to the high volume of imaging studies that need to be interpreted on a daily basis. Each study requires careful analysis and the generation of a detailed report that captures the relevant medical information accurately. One of the key benefits of automating the report generation process is the potential to alleviate the heavy workloads that radiologists encounter in clinical practice. The manual generation of reports is time-consuming and labor-intensive, often requiring radiologists to work under pressure to meet tight deadlines. This can lead to fatigue and an increased risk of errors, which may impact patient care and outcomes. By automating the report generation process, healthcare systems can reduce the time and effort required from radiologists, allowing them to focus on more complex cases and other critical tasks. This not only improves the efficiency and consistency of healthcare delivery but also enhances the overall quality of care by ensuring that radiologists can maintain a high level of accuracy and attention to detail in their work.

\subsection{Progress in Medical Report Generation}
The development of MRG methods has seen significant advancements over the years, driven by the integration of vision-and-language techniques~\cite{lin2023towards} that enable the automatic conversion of visual data from medical images into coherent textual descriptions. With the advent of deep learning, particularly the encoder-decoder framework, MRG methods have evolved to become more sophisticated and effective. The encoder-decoder framework forms the backbone of modern MRG systems. In this approach, the encoder processes the medical image, extracting meaningful features and representations, while the decoder generates the corresponding textual report. Therefore, researchers pay more attention to designing more powerful frameworks, especially the decoder module. Initially, decoders based on Long Short-Term Memory (LSTM)~\cite{hochreiter1997long} networks were widely used due to their ability to capture long-term dependencies and handle sequential data effectively~\cite{jing2017automatic,li2018hybrid}. As research in MRG progressed, the introduction of transformer-based architectures marked a significant leap forward~\cite{chen2020generating,li2021ffa}. Transformers~\cite{vaswani2017attention,han2024progressive}, with their attention mechanisms, allowed for better handling of long-range dependencies and provided a more nuanced understanding of the relationships between different parts of the input image and the generated text. This resulted in more accurate and contextually relevant reports, as transformers could dynamically focus on the most relevant regions of the image and generate text that reflected a deeper understanding of the medical content. The most recent advancements in MRG involve the utilization of large language models (LLMs)~\cite{wang2023r2gengpt,jin2024promptmrg,liu2024context}, which bring unprecedented capabilities in generating semantically coherent and contextually aware medical reports. LLMs, such as those based on architectures like GPT~\cite{openai2023chatgpt} or LLaMa~\cite{touvron2023llama}, leverage vast amounts of medical and general text data to learn the intricacies of language and clinical knowledge. These models can generate highly detailed and accurate reports by integrating contextual information from the medical images and aligning it with the rich, pre-learned knowledge from extensive textual corpora. The result is a level of report generation that closely mimics human expert interpretation, with the added benefit of consistency and efficiency.

\begin{figure}[h]
    \centering
    \includegraphics[width=1\linewidth]{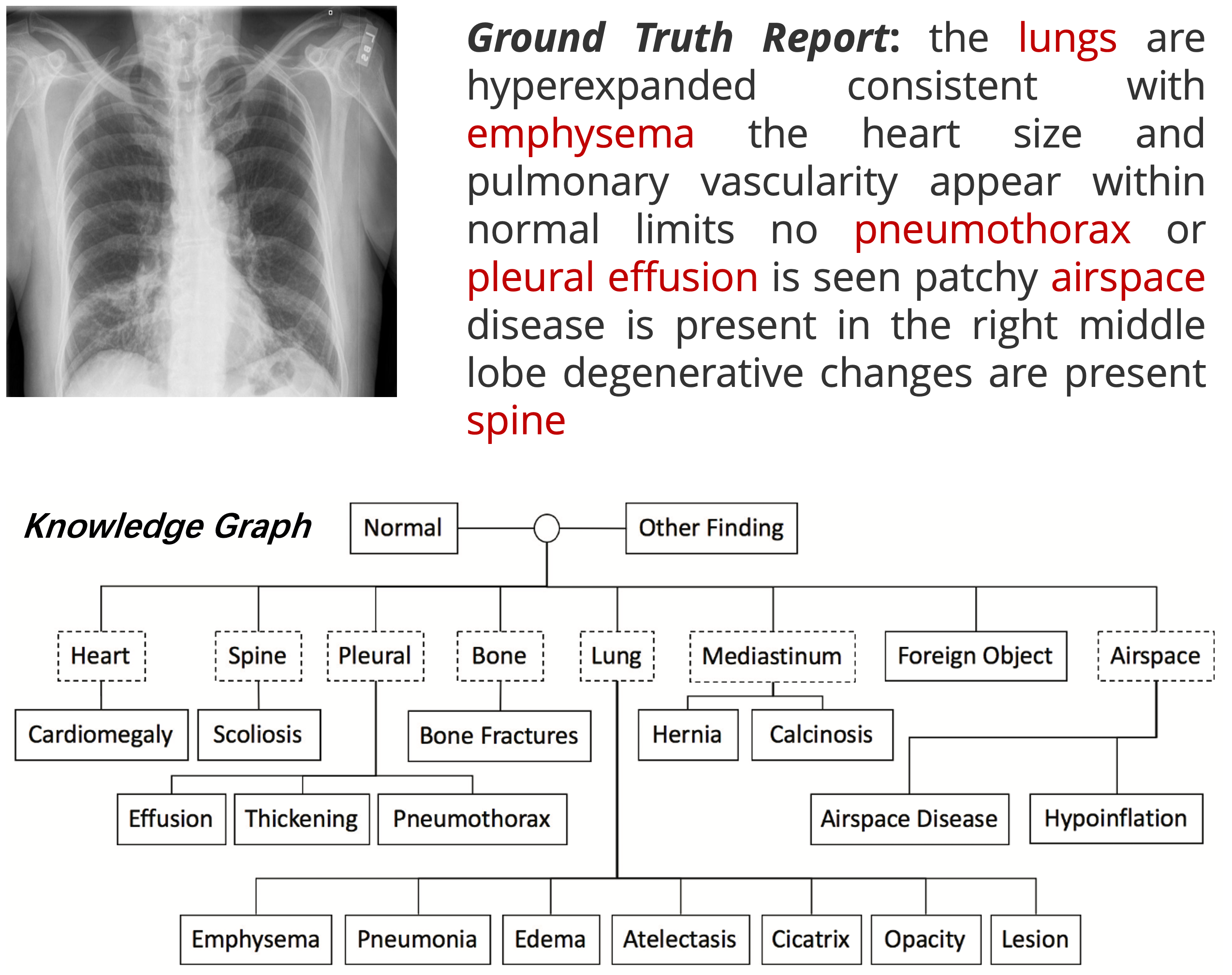}
    \caption{A sample from the IU X-ray dataset and the widely used knowledge graph. The red words are the nodes mentioned in the ground truth report.}
    \label{fig:mot}
\end{figure}

\subsection{Rethinking and New Perspective}

While existing approaches to MRG have focused on sequence generation using complex decoders, \textbf{\textit{we propose a new perspective that rethinks MRG as a multi-label classification problem}}. This perspective shift allows us to view the task of report generation not merely as creating coherent text from images but as identifying and classifying relevant medical concepts that can be directly used to construct reports. By treating MRG as a classification problem, we can simplify the task significantly and reduce the dependency on powerful and complex decoders, which are often necessary to generate detailed and semantically rich text. Instead, our approach focuses on accurately classifying keywords that are crucial for generating precise and relevant medical reports.

Our method leverages nodes from the commonly used knowledge graph~\cite{zhang2020radiology} as the primary keywords for classification. Knowledge graphs in the medical domain contain a wealth of structured information, including concepts, relationships, and clinical terminologies that are integral to medical reporting. The use of knowledge graphs in MRG is not a new concept~\cite{li2022cross,liu2021exploring,li2023dynamic,yang2022knowledge,li2023auxiliary,liang2024divide,li2024contrastive,li2023exploring}; they have been widely adopted due to their ability to encapsulate vast amounts of medical knowledge in an organized and easily accessible format. These graphs have been instrumental in enhancing the quality and depth of medical reports by providing a structured source of information that can be directly referenced during report generation. Researchers have continually enriched these graphs by adding more nodes and relationships to capture the complexity of medical knowledge. However, our research suggests that not all nodes in these graphs are equally useful for every specific case. Some nodes may be irrelevant or redundant, leading to unnecessary complexity and potential noise in the report generation process, as shown in Figure.\ref{fig:mot}. By extracting these nodes from existing reports, we can create a robust set of classification targets that cover the key aspects of medical findings. We integrate these classified nodes within a BLIP~\cite{li2022blip} framework, which facilitates the seamless integration of visual and textual data to generate medical reports. This approach allows us to generate reports by focusing on the classification of relevant keywords, which are then used to populate the report with accurate and clinically significant information.

\subsection{Contribution}

In this paper, we introduce a straightforward yet innovative framework that revolutionizes MRG by integrating classified keywords with a BLIP to produce detailed and accurate reports. Our primary argument is that by refining the existing knowledge graph and eliminating less relevant nodes, we can simplify MRG into a more manageable multi-label classification task. This approach focuses on identifying and classifying only the most pertinent nodes, which streamlines the report generation process and enhances the precision and clarity of the generated reports.

Our key contributions are as follows:
\begin{itemize}
    \item We propose a novel framework that can integrate classified keywords with BLIP for generating medical reports. This framework effectively utilizes a refined knowledge graph, focusing on the classification of key nodes to generate comprehensive reports. By simplifying the task to a multi-label classification problem, our framework reduces the reliance on complex decoders, enabling a more efficient and accurate report generation process. This represents a significant shift from traditional methods that rely on intricate sequence generation models.
    \item To validate our approach, we conduct extensive experiments on two benchmark datasets, namely IU X-ray~\cite{iuxray} and MIMIC-CXR~\cite{Johnson2019MIMICCXRAL}. These experiments demonstrate that refining the knowledge graph by identifying and removing less useful nodes simplifies the MRG task into a long-tailed multi-label classification problem. Our results show that this approach not only streamlines the report generation process but also significantly improves the accuracy and clarity of the reports. Compared to existing MRG systems, our framework achieves superior performance metrics, highlighting the effectiveness of our simplified approach.
    \item We believe that our perspective will serve as a foundational test bed for future research in the field of MRG. By demonstrating that MRG can be effectively simplified into a multi-label classification task, we encourage researchers to focus on the classification of essential concepts rather than on complex text generation. This shift in focus can lead to the development of more efficient and accurate MRG systems. Our framework offers a new direction for MRG research, underscoring the potential of integrating structured knowledge and classification techniques to advance the field.
    \item By focusing on the most relevant nodes within the knowledge graph, our framework ensures that the generated reports are not only accurate but also rich in clinically relevant information. This approach minimizes the inclusion of irrelevant or redundant information, leading to clearer and more precise reports that are better aligned with clinical needs.
\end{itemize}

\section{Related Work}

In this section, we discuss how existing medical report generation (MRG) systems utilize clinical-related keywords to improve the quality of generated reports. A significant line of research focuses on integrating knowledge graphs into MRG systems. Knowledge graphs provide a structured representation of medical knowledge, encompassing entities such as diseases, symptoms, treatments, and their interrelationships. This structured information serves as a valuable resource for enhancing the quality and coherence of automatically generated medical reports. Zhang \textit{et al.}~\cite{zhang2020radiology} proposed a radiology knowledge graph consisting of a root, 7 organ nodes, and 20 disease nodes. This reconstructed graph has been widely utilized in subsequent works to further enrich structured medical representations. For instance, Liu \textit{et al.}~\cite{liu2021exploring} retrieved semantically similar reports to enrich structured medical information, providing a more contextually relevant foundation for report generation. Building on this concept, both Li \textit{et al.}~\cite{li2023dynamic} and Yang \textit{et al.} extended this approach by extracting key entities from retrieved reports and converting them into triplets using RadGraph~\cite{jain2021radgraph}. Unlike these works that aim to enrich the reconstructed graph, our motivation to simplify MRG into a multi-label classification problem involves removing the irrelevant nodes for each specific case, thereby focusing on the most pertinent information. Another line of research focuses on utilizing clinical keywords. Initially, Jing \textit{et al.}~\cite{jing2017automatic} proposed the use of medical tags as topics to control sentence-level generation, ensuring that the generated text remained relevant to the clinical context. To harness the potential of large language models (LLMs), Jin \textit{et al.}~\cite{jin2024promptmrg} used diagnosis prompts to generate factually coherent reports, aligning the generated text closely with the diagnostic information. Following these concepts, we argue that accurate keywords (\textit{e.g.}, diseases, organs, or tissues) are critical in MRG tasks. To obtain these keywords, a classification framework can be employed, which allows for precise identification and use of relevant clinical terms in report generation.

These approaches demonstrate the evolving landscape of MRG systems and highlight the critical role of structured knowledge and clinical keywords in improving the quality and accuracy of generated medical reports. Our work builds on these insights by simplifying the task into a multi-label classification problem, focusing on the classification of key medical concepts to generate more streamlined and accurate reports.

\section{Method}

\begin{figure*}
    \centering
    \includegraphics[width=0.8\linewidth]{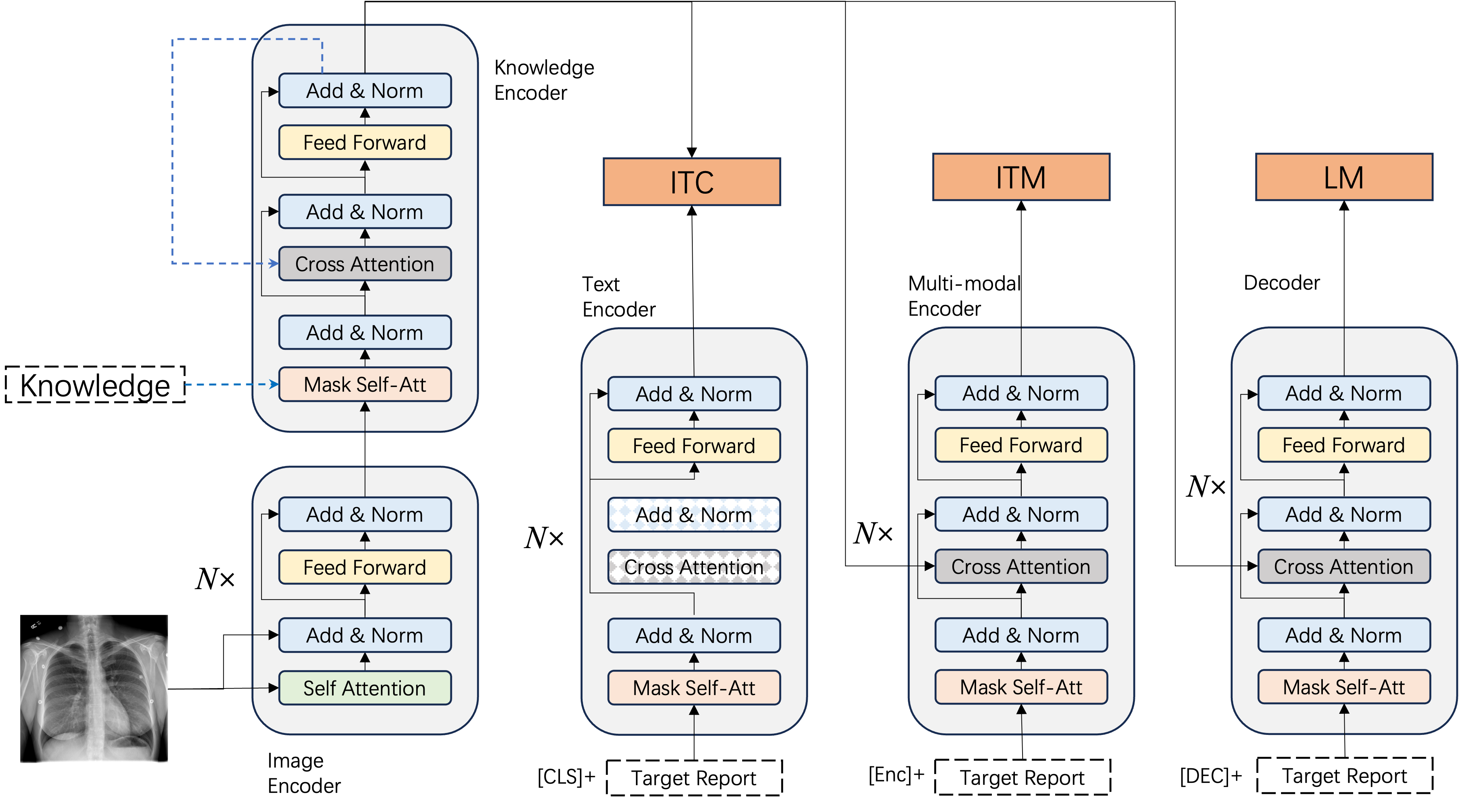}
    \caption{The overview of our proposed framework, which consists of a knowledge encoder, image encoder, text encoder, multi-modal encoder, and a decoder. The knowledge encoder integrates classified nodes with visual representations.}
    \label{fig:overview}
\end{figure*}

In this section, we introduce the detailed implementation of our proposed method, which integrates a BLIP~\cite{li2022blip} and a knowledge encoder to enhance the quality of medical report generation. The overall architecture of our model is depicted in Figure.\ref{fig:overview}.

\subsection{Model Architecture}

The BLIP is central to our model and facilitates the fusion of visual and textual data for report generation. BLIP combines features from different modalities through cross-modal attention mechanisms, ensuring coherent and contextually relevant report generation.

\textbf{Image Encoder:} The image encoder in our framework employs a ViT-L~\cite{vit} architecture. This encoder is pre-trained on an extensive dataset of image-text pairs. Input images are divided into 196 patches, each of which is processed through the encoder along with a [CLS] token added to the beginning. The core mechanism of the ViT-L is the self-attention mechanism, defined as:

\begin{equation}
\text{Attention}(Q, K, V) = \text{softmax}\left(\frac{QK^T}{\sqrt{d_k}}\right)V
\end{equation}

where $Q$, $K$, and $V$ represent the query, key, and value matrices derived from the visual tokens. The final output is encoded into visual vectors $\mathbf{f}_I$, which are used in the subsequent report generation stages.

\textbf{Text Encoder:} The textual encoder in our framework is based on BERT~\cite{bert}. BERT processes text bidirectionally, integrating context from both directions to provide a rich understanding of the input text. The text encoder transforms the input text into a sequence of hidden states $\mathbf{h}_T$, which capture the semantic information required for report generation.

\textbf{Knowledge Encoder:} The knowledge encoder, also based on BERT, processes clinical-related keywords classified from the given image. The parameters in the knowledge encoder, except for those in the attention layers, are shared with the text encoder to ensure that features are mapped to the same space. The knowledge encoder outputs hidden representations $\mathbf{h}_K$ for the classified nodes, representing the relevant medical concepts.

\textbf{Cross-Attention Mechanism:} To merge features from different modalities, we utilize a cross-attention mechanism. The query ($Q$) is derived from the visual representations $\mathbf{f}_I$, while the key ($K$) and value ($V$) are derived from the node representations $\mathbf{h}_K$ obtained from the knowledge encoder. The cross-attention mechanism is defined as:

\begin{equation}
\text{CrossAttention}(Q, K, V) = \text{softmax}\left(\frac{QK^T}{\sqrt{d_k}}\right)V
\end{equation}

This mechanism integrates the visual and node representations, enhancing the image features with structured medical knowledge for more effective report generation.

\textbf{Text Decoder:} The text decoder in our model transforms the encoded features into natural language. It replaces the self-attention layers of the text encoder with causal self-attention layers and a feedforward network, enabling the generation of coherent and contextually relevant text. The text decoder utilizes the enhanced image features to generate the final report.

\subsection{Training Objective}

In this section, we introduce the loss functions used in model training. It is important to note that the classifier (a ResNet-50~\cite{he2016deep}) is trained separately.

\textbf{Image-Text Contrastive Loss:} This loss function aims to align the feature spaces of the enhanced image features and the text features extracted by the text encoder. Using contrastive learning, it encourages positive image-text pairs to have similar representations, while negative pairs are pushed apart. This loss is effective in improving vision and language understanding~\cite{radford2021learning, li2021align}. Our implementation follows the approach by Li \textit{et al.}~\cite{li2023dynamic}, introducing a momentum encoder to produce features.

\begin{equation}
\mathcal{L}_{\text{contrastive}} = -\log\frac{\exp(\text{sim}(\mathbf{f}_I, \mathbf{h}_T) / \tau)}{\sum_{j} \exp(\text{sim}(\mathbf{f}_I, \mathbf{h}_T^j) / \tau)}
\end{equation}

where $\text{sim}(\cdot)$ denotes the similarity function and $\tau$ is a temperature parameter.

\textbf{Image-Text Matching Loss:} This loss function activates the text encoder, helping the model learn image-text multimodal representations and align them effectively.

\begin{equation}
\mathcal{L}_{\text{matching}} = -\log P(y=1 | \mathbf{f}_I, \mathbf{h}_T)
\end{equation}

\textbf{Language Modeling Loss:} The language modeling loss activates the text decoder and aims to help the model learn to generate the report based on the given features. It optimizes the cross-entropy loss and trains the model to maximize the likelihood of the text in an autoregressive manner, enabling the model with generation capabilities.

\begin{equation}
\mathcal{L}_{\text{LM}} = -\sum_{t=1}^T \log P(w_t | w_{1:t-1}, \mathbf{f}_I, \mathbf{h}_K)
\end{equation}

To improve efficiency, the parameters in the self-attention layers of the text encoder and text decoder are shared, as these layers effectively capture the differences between encoding and decoding. And the total loss objective in this work is the sum of these three objectives.

\begin{table*}[t]
\centering
\caption{Performance Comparison on IU-Xray and MIMIC-CXR Datasets}
\label{tab:performance}
\resizebox{0.8\linewidth}{!}{\begin{tabular}{c|c|c|c|c|c|c|c|c|c}
\toprule
Dataset   & Methods        & Year                 & BLEU-1  & BLEU-2  & BLEU-3  & BLEU-4  & ROUGE   & METEOR  & CIDEr   \\ \hline
          & Transformer\cite{vaswani2017attention}    & 2017             & 0.372   & 0.251   & 0.147   & 0.136   & 0.317   & 0.168   & 0.310   \\ \cline{2-10} 
          & M2transformer\cite{cornia2020meshed}  & 2020             & 0.402   & 0.284   & 0.168   & 0.143   & 0.328   & 0.170   & 0.332   \\ \cline{2-10} 
          & R2Gen\cite{chen2020generating}          & 2020            & 0.470   & 0.304   & 0.219   & 0.165   & 0.371   & 0.187   & -       \\ \cline{2-10} 
          & R2GenCMN\cite{chen-acl-2021-r2gencmn}       & 2021              & 0.475   & 0.309   & 0.222   & 0.170   & 0.375   & 0.191   & -       \\ \cline{2-10} 
        & MSAT\cite{wang2022medical}           & 2022           & 0.481   & 0.316   & 0.226   & 0.171   & 0.372   & 0.190   & 0.394   \\ \cline{2-10} 
 IU-Xray       & METransformer\cite{wang2023metransformer}  & 2023             & 0.483   & 0.322   & 0.228   & 0.172   & 0.380   & 0.192   & 0.435   \\ \cline{2-10} 
          & R2GenGPT(Deep)\cite{wang2023r2gengpt} & 2023   & 0.488   & 0.316   & 0.228   & 0.173   & 0.377   & 0.211   & 0.438   \\ \cline{2-10} 
          & DCL \cite{li2023dynamic}           & 2023             & 0.468       & 0.311       & 0.237       & 0.163   & 0.383   & 0.193   & 0.586   \\ \cline{2-10} 
          & KiUT \cite{huang2023kiut}          & 2023             & 0.525   & 0.360   & 0.251   & 0.185   & 0.409   & 0.242   & -       \\ \cline{2-10} 
          & PromptMRG\cite{jin2024promptmrg}     & 2024             & 0.401   & -       & -       & 0.098   & 0.281   & 0.160   & -       \\ \cline{2-10} 
          & InVERGe\cite{deria2024inverge}       & 2024             & \textbf{0.499} & 0.324   & 0.226   & 0.168   & 0.384   & 0.194   & -       \\ \cline{2-10} 
          & Ours(100\%)    &                      & 0.494   & \textbf{0.389} & \textbf{0.330} & \textbf{0.292} & \textbf{0.480} & \textbf{0.265} & \textbf{2.229} \\ \hline
 
          & Transformer\cite{vaswani2017attention}    & 2017             & 0.316   & 0.199   & 0.140   & 0.092   & 0.267   & 0.129   & 0.134   \\ \cline{2-10} 
          & M2transformer\cite{cornia2020meshed}  & 2020             & 0.332   & 0.210   & 0.142   & 0.101   & 0.264   & 0.134   & 0.142   \\ \cline{2-10} 
          & R2Gen\cite{chen2020generating}          & 2020            & 0.353   & 0.218   & 0.145   & 0.103   & 0.277   & 0.142   & -       \\ \cline{2-10} 
          & R2GenCMN\cite{chen-acl-2021-r2gencmn}       & 2021              & 0.353   & 0.218   & 0.148   & 0.106   & 0.278   & 0.142   & -       \\ \cline{2-10} 
  & PPKED\cite{liu2021exploring}          & 2021             & 0.360   & 0.224   & 0.149   & 0.106   & 0.284   & 0.149   & 0.237   \\ \cline{2-10} 
          & MSAT\cite{wang2022medical}           & 2022           & 0.373   & 0.235   & 0.162   & 0.120   & 0.282   & 0.143   & 0.299   \\ \cline{2-10} 
          & GSK\cite{yang2022knowledge}           & 2022 & 0.363 & 0.228   & 0.156   & 0.115   & 0.284   & -       & 0.203   \\ \cline{2-10} 
 MIMIC-CXR  & R2GenGPT(Deep)\cite{wang2023r2gengpt} & 2023   & 0.411   & \textbf{0.267} & \textbf{0.186} & 0.134   & 0.297   & 0.160   & 0.269   \\ \cline{2-10} 
          & METransformer\cite{wang2023metransformer}  & 2023             & 0.386   & 0.250   & 0.169   & 0.124   & 0.291   & 0.152   & \textbf{0.362} \\ \cline{2-10} 
          
          & DCL\cite{li2023dynamic}           & 2023             & 0.370       & 0.231       & 0.154       & 0.109   & 0.284   & 0.150   & 0.281   \\ \cline{2-10} 
          & KiUT\cite{huang2023kiut}           & 2023             & 0.393   & 0.243   & 0.159   & 0.113   & 0.285   & 0.160   & -       \\ \cline{2-10} 
          & RGRG\cite{tanida2023interactive}           & 2023             & 0.373   & -       & -       & 0.126   & 0.264   & 0.168   & -       \\ \cline{2-10} 
          & PromptMRG\cite{jin2024promptmrg}      & 2024             & 0.398   & -       & -       & 0.112   & 0.268   & 0.157   & -       \\ \cline{2-10} 
          & InVERGe\cite{deria2024inverge}        & 2024             & \textbf{0.425} & 0.240   & 0.132   & 0.100   & \textbf{0.309} & \textbf{0.175} & -       \\ \cline{2-10} 
          & Ours(100\%)    &                      & 0.347   & 0.234   & 0.172   & \textbf{0.134} & 0.297   & 0.170   & 0.305   \\ \bottomrule
\end{tabular}}
\end{table*}

\section{Experiment}
\subsection{Setting}

\noindent\textbf{Datasets}
We trained and tested our model on two commonly used datasets for medical report generation tasks, namely IU-Xray\cite{iuxray} and MIMIC-CXR\cite{Johnson2019MIMICCXRAL}. Following the previous settings\cite{chen2020generating,li2023dynamic}, we adopted the same preprocess for these two datasets.

IU-Xray is extensively utilized to assess the performance of radiology reporting systems. This dataset comprises 3,955 radiology reports and 7,470 chest X-ray images, each linked to frontal or both frontal and lateral view images. In line with previous techniques, we also excluded cases that contained only a single image, leaving us with 2,069 cases for training, 296 for validation, and 590 for testing. MIMIC-CXR is a comprehensive publicly available dataset of chest radiographs accompanied by free-text radiology reports. It includes 10 folders containing a total of 377,110 chest X-ray images and 227,835 corresponding reports. For this experiment, we utilized the same version of the pre-processed dataset as used in previous methods.

\noindent\textbf{Metrics}
To evaluate the quality of the generated reports, we employed four commonly used evaluation metrics: BLEU\cite{papineni-etal-2002-bleu}, METEOR\cite{banerjee2005meteor}, ROUGE-L\cite{lin-2004-rouge}, and CIDEr\cite{vedantam2015cider}. BLEU measures the precision of n-grams in the generated text compared to reference texts, emphasizing accuracy and fluency. However, due to the repetitive pasting of the same text, BLEU may not accurately reflect the quality of the reports. CIDEr, on the other hand, better rewards topic-related vocabulary and penalizes redundant and repetitive words. Additionally, ROUGE-L and METEOR are also commonly used evaluation metrics, evaluating the alignment between generated and reference texts.

\noindent\textbf{Experimental Details} For our IU-Xray dataset, we trained our model on 8 NVIDIA 3090 GPUs
with batch sizes 2 and 30 epochs. For MIMIC-CXR dataset, we trained our model on 2 NVIDIA A6000 GPUs with batch sizes 16 and 30 epochs. The checkpoint acquires
the highest BLEU4 metric is saved and used for testing. We utilized a pretrained Visual Transformer (ViT) as image feature extractor with 768 visual width. The learning
rate is set as 1e-4 and the optimizer is AdamW [34] with a
weight decay of 5e-5. For the report generation task, we used the Blip decoder model with the bert-base tokenizer. The model architecture is based on BERT with a hidden size of 768, 12 hidden layers, and 12 attention heads. We used a dropout probability of 0.1 and the GELU activation function. The vocabulary size was 30,524 tokens, and the maximum position embedding size was 512. Additionally, cross-attention was enabled to introduce external node knowledge.

To evaluate the model's performance under different classification accuracy levels, we randomly masked and added noise to original node labels and created node knowledge with mean accuracy levels of 70\%, 80\%, and 90\%, respectively. We then trained the report generation model with these imperfect classification results. 

\noindent\textbf{Baseline} In our performance comparison, we selected several well-established baselines to evaluate the effectiveness of our approach on the IU-Xray and MIMIC-CXR datasets. These baselines include Transformer-based models such as the original Transformer model \cite{vaswani2017attention} and the M2Transformer \cite{cornia2020meshed}, which have demonstrated strong performance in image captioning and report generation tasks. Additionally, we included R2Gen \cite{chen2020generating} and its enhanced versions R2GenCMN \cite{chen-acl-2021-r2gencmn} and R2GenGPT(Deep) \cite{wang2023r2gengpt}, which are specifically designed for medical report generation. Recent state-of-the-art methods such as MSAT \cite{wang2022medical}, METransformer \cite{wang2023metransformer}, and dynamic models like DCL \cite{li2023dynamic} and KiUT \cite{huang2023kiut} which incorporate knowledge were also incorporated to provide a comprehensive evaluation. We also included recent We also included the latest report generation models, such as PromptMRG\cite{jin2024promptmrg} and InVERGe\cite{deria2024inverge}. These baselines were chosen due to their relevance and high performance in similar tasks, allowing us to rigorously compare our proposed method's performance across multiple metrics, including BLEU, ROUGE, METEOR, and CIDEr.

\subsection{Main result}
Table \ref{tab:performance} presents a comprehensive performance comparison of our proposed method against several state-of-the-art models on the IU-Xray and MIMIC-CXR datasets.
Our model denoted as Ours(100\%), achieves superior performance on several metrics across both datasets. Specifically, our model achieves the highest BLEU-4 score of 0.292 on the IU-Xray dataset, surpassing the best score of 0.185 by KiUT \cite{huang2023kiut}. Furthermore, our model also leads in METEOR and CIDEr scores with values of 0.265 and 2.229, respectively. These results demonstrate the robustness and effectiveness of our approach in generating high-quality medical reports.
Comparatively, the METransformer \cite{wang2023metransformer} and R2GenGPT(Deep) \cite{wang2023r2gengpt} also perform well, indicating significant advancements in the field. However, our model consistently outperforms these methods, particularly in the CIDEr metric, which is critical for evaluating the informativeness of the generated reports. These results suggest that our approach captures the necessary details and provides a more comprehensive and contextually accurate summary of the medical findings.

The results on the MIMIC-CXR dataset further validate our model's superiority. Notably, our model and R2GenGPT \cite{wang2023r2gengpt} both achieve a BLEU-4 score of 0.134, but our model significantly outperforms R2GenGPT in the CIDEr metric, achieving a score of 0.305 compared to R2GenGPT's 0.269. This indicates that our model excels in generating reports that are not only syntactically accurate but also rich in content, providing detailed and relevant medical information. This advantage is attributed to incorporating knowledge nodes, which enhance the model's ability to generate medically relevant vocabulary. Moreover, considering that the R2GenGPT model utilizes the more powerful GPT architecture for decoding, our results are awe-inspiring and could be further improved with the more powerful decoder.
Comparatively, our model also surpasses other state-of-the-art methods. For instance, while the METransformer \cite{wang2023metransformer} achieves a slightly higher CIDEr score of 0.362, our model demonstrates competitive performance across all metrics, including BLEU-1, BLEU-2, BLEU-3, and METEOR. This consistent high performance highlights the generalizability and reliability of our model in diverse clinical scenarios. The ability to maintain high performance in both IU-Xray and MIMIC-CXR datasets showcases the model's adaptability and potential for broader application in medical report generation tasks.

\begin{figure}[htbp]
    \centering
    \includegraphics[width=\linewidth]{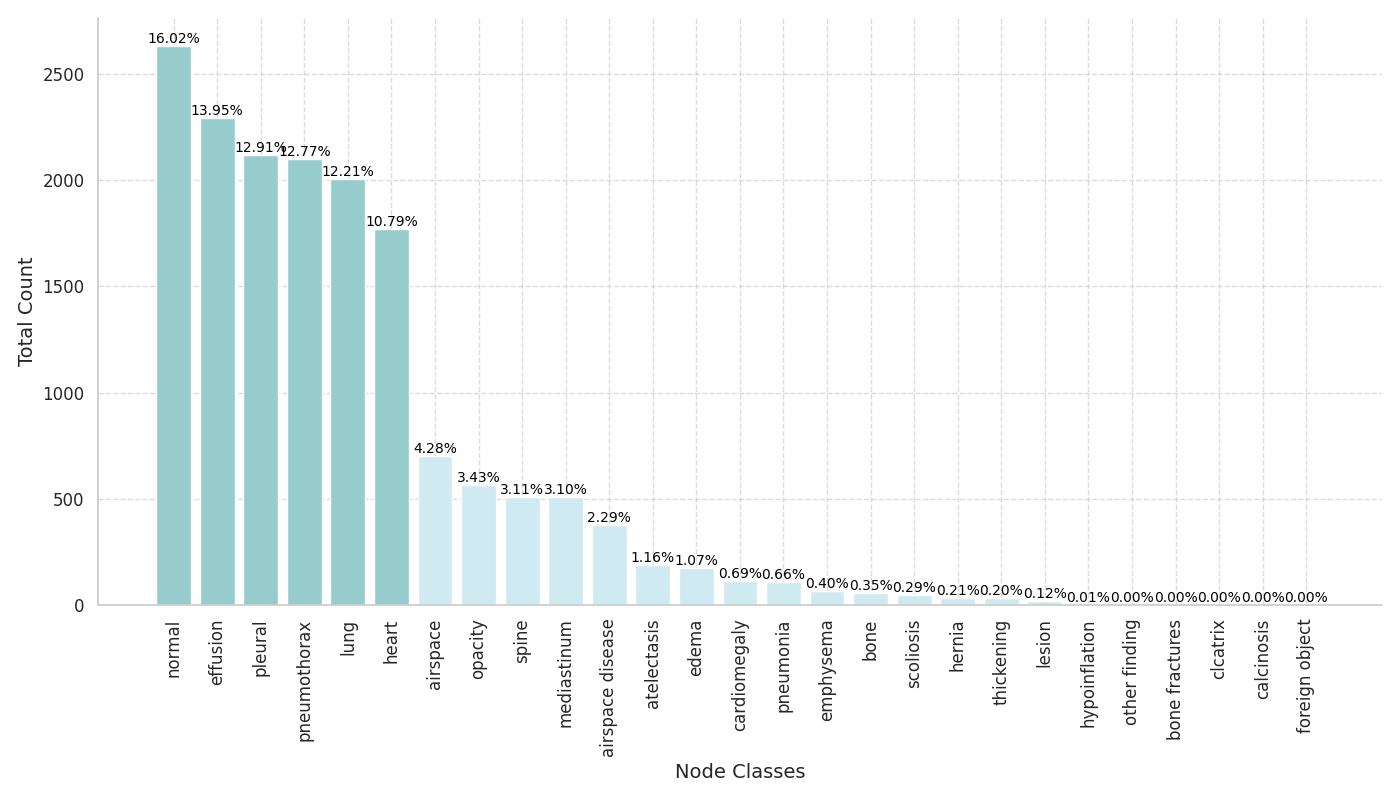} 
    \caption{Label distribution in the IU-Xray dataset.}
    \label{fig:iu_dis}
\end{figure}
\begin{figure}[htbp]
    \centering
    \includegraphics[width=\linewidth]{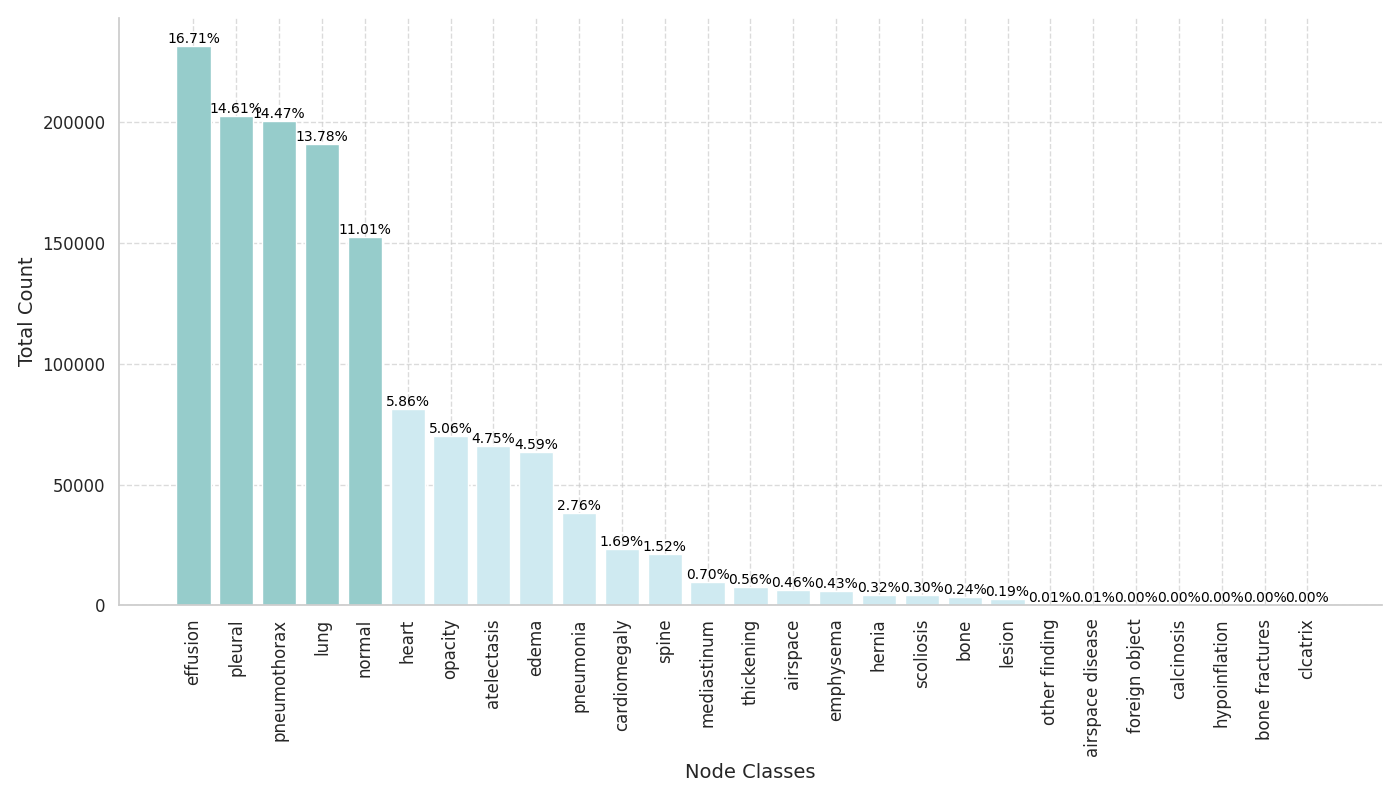} 
    \caption{Label distribution in the MIMIC-CXR dataset.}
    \label{fig:mimic_dis}
\end{figure}

\begin{figure}[htbp]
    \centering
    \includegraphics[width=\linewidth]{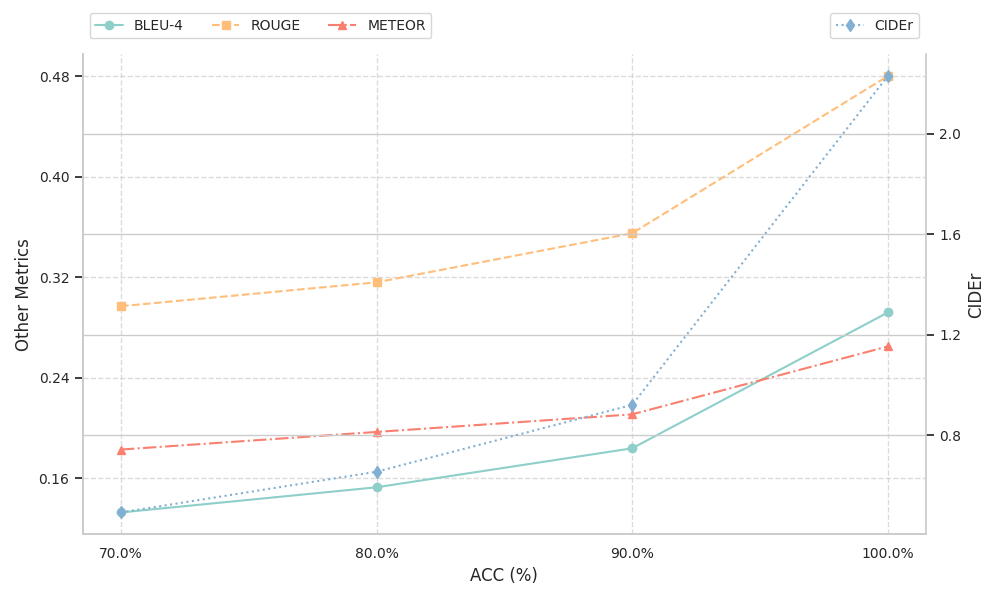} 
    \caption{Changes on natural general metrics along with node accuracy on the IU-Xray dataset.}
    \label{fig:iu_xray}
\end{figure}

\begin{figure}[htbp]
    \centering
    \includegraphics[width=\linewidth]{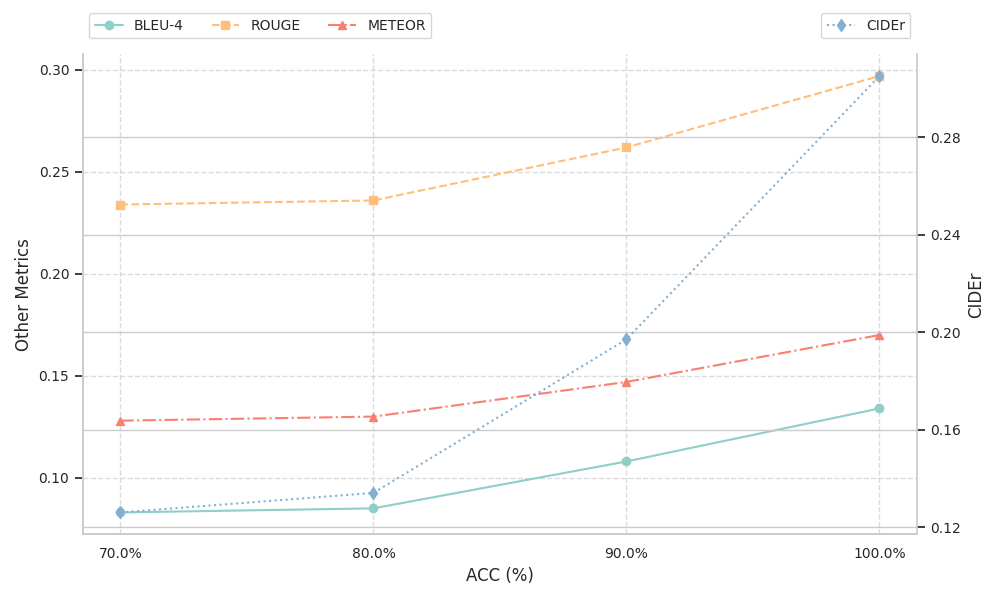} 
    \caption{Changes on natural general metrics along with node accuracy on the MIMIC dataset.}
    \label{fig:mimic}
\end{figure}

\begin{figure*}[t]
    \centering
    \includegraphics[width=\linewidth]{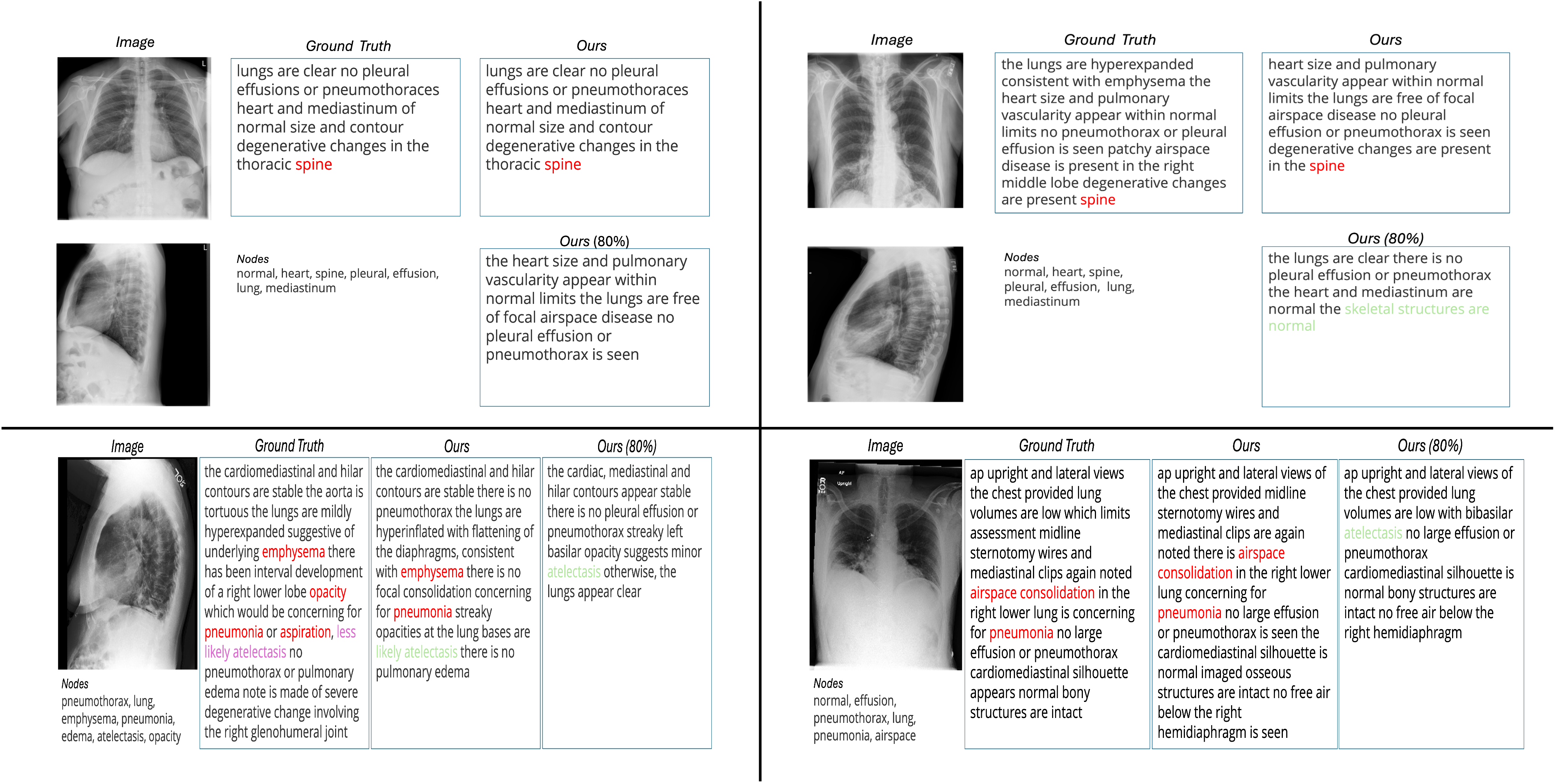} 
    \caption{Illustrations of samples, including ground truth and predictions, from both IU-Xray and MIMIC datasets.}
    \label{fig:case}
\end{figure*}
\subsection{Discussion}

\noindent\textbf{Why ResNet Does Not Work}

The distribution of nodes in both the IU-Xray and MIMIC-CXR datasets reveals a classic long-tailed classification problem, which significantly impedes the performance of classifiers. As shown in Fig.~\ref{fig:iu_dis}, the IU-Xray dataset has a significant imbalance, with a few classes like "normal", "effusion", "pleural", and "lung" dominating the distribution, accounting for a large portion of the total instances. In contrast, many other classes, such as "hypoinflation", "lesion", and "foreign object", are represented by very few instances. This imbalance causes the model to skew predictions towards these dominant classes to achieve higher average accuracy. However, this behaviour suppresses the learning and representation of other vital labels, especially critical disease tags. As a result, rare conditions are often not predicted accurately. In the IU-Xray dataset, this imbalance is more pronounced because many label combinations appear only once or twice in the entire dataset, making it difficult for the model to learn and classify these features accurately. Fig.~\ref{fig:mimic_dis} displays the distribution for the MIMIC-CXR dataset, which similarly shows a heavy imbalance. Despite the larger dataset and more samples, the MIMIC-CXR dataset still needs a severe long-tailed problem. Classes like "effusion", "pleural", and "pneumothorax" have significantly more instances compared to classes such as "calcinosis", "hypoinflation", and "bone fractures." This imbalance affects the model's performance similarly, leading it to prioritize normal classes and overlook rare but essential conditions.

As shown in Table \ref{tab:multilabel_performance}, despite achieving high accuracy (aACC) values of 0.893 for IU-Xray and 0.901 for MIMIC-CXR, other metrics such as aAUC, aF1, and mAP are less impressive. This discrepancy occurs because the model predicts the dominant classes to boost overall accuracy more frequently, adversely affecting its performance on less frequent but critical labels. Specifically, the aAUC values of 0.407 for IU-Xray and 0.489 for MIMIC-CXR, and the aF1 scores of 0.186 and 0.211, respectively, highlight the model's inadequate discriminative power and balance between precision and recall across all classes. Additionally, the mAP values of 0.206 for IU-Xray and 0.209 for MIMIC-CXR further emphasize the model's struggle to maintain precision across diverse labels.

While ResNet-50 can capture the more common patterns in the datasets, it could generalize better to the less frequent conditions, which are crucial for comprehensive medical diagnosis. Utilizing predicted nodes based on this classification further compounds these issues, as the downstream report generation heavily relies on the accuracy and completeness of these predictions. Therefore, addressing the long-tailed distribution through more sophisticated techniques is necessary to improve the performance of the report generation task.
\begin{table}[htbp]
\centering
\caption{Multi-Label Classification Performance on IU-Xray and MIMIC-CXR Datasets}
\label{tab:multilabel_performance}
\begin{tabular}{lcccc}
\toprule
Dataset    & aAUC & aF1  & aACC  & mAP  \\ 
\midrule
IU-Xray    & 0.407 & 0.186 & 0.893 & 0.206 \\ 
MIMIC-CXR  & 0.489 & 0.211 & 0.901 & 0.209 \\ 
\bottomrule
\end{tabular}
\end{table}

\noindent\textbf{How Does Node Accuracy Affect the Performance}
Fig.~\ref{fig:iu_xray} demonstrates the performance trends of our report generation model on the IU-Xray dataset under varying classification accuracy levels (70\%, 80\%, 90\%, and 100\%). All evaluated metrics exhibit notable improvements with increasing accuracy. Specifically, BLEU-4 rises from 0.133 at 70\% to 0.292 at 100\% accuracy, ROUGE scores increase from 0.297 to 0.480, and METEOR improves from 0.183 to 0.265. These enhancements suggest that higher classification accuracy positively influences the quality of generated reports, as correctly identified nodes contribute valuable knowledge to the report generation process, thereby boosting metrics that depend on content overlap.

Moreover, the CIDEr metric shows a marked improvement, climbing from 0.49 to 2.229 as accuracy reaches 100\%. This substantial increase in CIDEr underscores the model's enhanced ability to generate detailed and contextually relevant reports, significantly when the classification of rare conditions improves. The long-tailed distribution of the IU-Xray dataset means that lower accuracy levels result in the over-prediction of common conditions, which hampers the CIDEr score. In contrast, near-perfect accuracy facilitates the detection of rare abnormalities, thus maximizing the CIDEr metric. These results highlight the critical role of accurate classification in achieving superior report generation quality. 

Fig.~\ref{fig:mimic} demonstrates that similar to the IU-Xray dataset, the evaluation metrics for report generation on the MIMIC-CXR dataset consistently improve as classifier accuracy increases. For instance, BLEU-4 rises from 0.083 at 70\% accuracy to 0.134 at 100\% accuracy, ROUGE scores increase from 0.234 to 0.297, and METEOR improves from 0.128 to 0.170. These trends indicate that continually enhancing classifier accuracy and integrating multi-label classification results into the decoder can significantly boost model performance.

\noindent\textbf{Case Study}
To further investigate the effectiveness of our method, we conducted a qualitative analysis on both the IU-Xray and MIMIC-CXR datasets, utilizing node knowledge and comparing reports generated by our model with the ground truth. To emphasize the importance of the classifier, we also evaluated reports generated using classification results with 80\% accuracy. 

The upper half of Fig.~\ref{fig:case} pertains to the IU-Xray dataset. In the top-left image, our model demonstrates accurate predictions due to the incorporation of relevant knowledge nodes. The model successfully identifies degenerative changes in the spine and accurately predicts keywords such as "pleural". However, when the classifier accuracy is reduced to 80\%, the model fails to detect issues in the spine. In the top-right image, our model again shows excellent performance. It accurately identifies abnormalities in the spine, although it misses the "airspace disease" anomaly. In contrast, the model with 80\% classification accuracy incorrectly predicts the bone structure in the spine as normal. It is worth noting that IU-Xray is a highly imbalanced dataset with very few disease samples, causing both the classifier and the generator to be biased towards "normal" or "no disease." Therefore, accurately describing abnormalities in the reports is particularly challenging and a significant achievement. 

The lower half of Fig.~\ref{fig:case} corresponds to the MIMIC-CXR dataset. In the bottom-left image, our model demonstrates its capability by accurately predicting the presence of diseases such as "emphysema", "pneumonia", and "opacity" while missing "aspiration" and incorrectly predicting "atelectasis". The model with 80\% classification accuracy also incorrectly predicts "atelectasis", marked in green. Such failure highlights a limitation of our model: its reliance on predicting whether an image is related to a knowledge node rather than directly predicting abnormalities, thereby affecting its disease detection accuracy. Similarly, in the bottom-right image, our model accurately predicts diseases, whereas the 80\% accuracy model fails to do so.

Therefore, we can see that the model's effectiveness is highly dependent on the accuracy of the classification results. The more accurate the classification, the better the report can reflect the presence of diseases. However, this approach also has limitations, as it does not effectively determine the presence of diseases. The classifier only predicts whether the related nodes appear in the report, highlighting areas for future improvement.

\noindent\textbf{Limitation} While we proposed the idea of simplifying MRG into a multi-label classification problem and validated it with ground truth data, we did not fully tackle the long-tailed classification issue inherent in this task. In medical contexts, some classes (e.g., common diseases) appear frequently in training data, while others (e.g., rare conditions) are underrepresented. This imbalance can lead to biased models that perform well on common classes but poorly on rare ones, which can be crucial in clinical decision-making where detecting rare but critical conditions is often necessary.

\section{Conclusion}
In this paper, we introduced a novel perspective on MRG, proposing that it can be simplified into a multi-label classification problem. By utilizing 27 nodes from a knowledge graph as targets for multi-label classification, we developed an innovative framework that integrates these nodes with a BLIP. Our approach challenges the conventional reliance on complex decoders, demonstrating that even with a simpler decoder, the accurate identification of key nodes can lead to the generation of high-quality medical reports. The results from our extensive experiments clearly indicate that our framework not only simplifies the MRG task but also achieves SOTA performance across various metrics. We hope that our work inspires other researchers in the field to explore the application of long-tailed classification techniques within MRG systems. By doing so, they may uncover more effective and efficient methods for tackling the challenges of medical report generation, ultimately contributing to the advancement of automated medical diagnostics and the improvement of healthcare outcomes.

\bibliographystyle{IEEEtran}
\bibliography{references}

\end{document}